\documentclass{article}
\usepackage{amsmath,amssymb}

\usepackage[final,dblblindworkshop]{neurips_2026}
\workshoptitle{AI for Accelerated Materials Discovery (AI4Mat)}

\usepackage[utf8]{inputenc}
\usepackage{pdfpages}

\usepackage[T1]{fontenc}
\usepackage{hyperref}
\usepackage{url}
\usepackage{booktabs}
\usepackage{graphicx}
\usepackage{amsmath}
\usepackage{amssymb}
\usepackage{microtype}
\usepackage{xcolor}
\usepackage[labelfont=bf]{caption}

\newcommand{\rev}[1]{#1} 
\newenvironment{revblock}{}{}

\newcommand{\tbv}[1]{#1}

\newcommand{\crfix}[1]{#1}
\newenvironment{crblock}{}{}

\title{MatHarness: Towards Long-Horizon Open-ended AutoResearch for Computational Materials Science}

\author{%
  Yanlin Fei\textsuperscript{1}\thanks{Equal contribution.} \quad
  Lei Li\textsuperscript{1}\footnotemark[1] \quad
  Shaolong Chen\textsuperscript{1} \quad
  Nazhou Liu\textsuperscript{1} \quad
  Xinmiao Yu\textsuperscript{1} \\
  \textbf{Madalina Ciobanu}\textsuperscript{1} \quad
  \textbf{Navan Preet Singh}\textsuperscript{1} \quad
  \textbf{Qingqing Mao}\textsuperscript{1,2}%
  \thanks{Corresponding author.} \quad
  \textbf{Ritankar Das}\textsuperscript{1,2} \\[0.6em]
  \textsuperscript{1}Prentis AI, San Francisco, CA \\
  \textsuperscript{2}Titan Holdings, San Francisco, CA \\[0.3em]
  \texttt{qmao@prentis.ai}
}

\begin{document}

\maketitle

\begin{abstract}
Materials science research is a non-linear, long-horizon process. Existing scientific agents, however, largely rely on specific task execution or predefined workflows, rendering them brittle in open-ended settings. To bridge this gap, we present MatHarness, an agent harness designed for open-ended, long-horizon computational materials research spanning the complete lifecycle from initial ideation to manuscript writing. MatHarness couples three mechanisms: execution robustness through pre-flight verification, expert-distilled skills, and just-in-time auditing; epistemic adaptability through persistent research artifacts and an evolving scientific mental model; and collective intelligence through a heterogeneous multi-agent round table under targeted human oversight. We evaluate MatHarness on adsorbate-induced surface segregation in Cu--Zn alloys, an open question in methanol-synthesis catalysis. Over a multi-week trajectory consuming roughly $10^3$ GPU-hours---50 production DFT relaxations prescreened by 171 machine-learning-potential evaluations---the system revised its plan six times, recovered autonomously from failed calculations, and had an independent auditor overturn one of its own over-attributed claims before publication. Within the sampled metallic dilute-limit models, the validated DFT analysis supports an adsorbate-induced tendency toward greater surface Cu exposure at both composition endpoints, with facet- and coverage-dependent exceptions. These results were compiled into a complete manuscript with human involvement concentrated in specification, approval, and strategic steering gates. The code is publicly available at \url{https://github.com/PrentisAI/pippin}.
\end{abstract}

\section{Introduction}
\label{sec:introduction}

Computational materials research is an iterative, long-horizon decision process. Rather than executing isolated tasks, researchers must continuously hypothesize mechanisms, construct valid  models, troubleshoot non-convergent simulations, and interpret intermediate outputs to determine the next informative experiment. These decisions are deeply coupled, and navigating the vast space of hypotheses and experiments demands continuous, context-aware reasoning in which early mistakes compound over time.

Computational materials science is an ideal candidate for the emerging AutoResearch paradigm: its workflows and artifacts are natively digital, while the heavy cost of high-fidelity simulations makes rigorous research planning, robust execution, and adaptive experiment selection high-leverage.

Recent scientific agents have evolved from tool-augmented systems~\citep{bran2024chemcrow,boiko2023coscientist} and domain-specific task automation~\citep{ghafarollahi2025atomagents,chaudhari2026matsciagent,ock2026adsorbagent} to long-horizon execution~\citep{vriza2026atomisticsim,qian2026autosci,zhao2026scienceflow}. Despite this progress, existing agentic systems are mainly built around linear task-solving or predefined workflows. This paradigm fails to capture the complexity of real-world computational materials research, where simulations frequently fail to converge, intermediate feedback invalidates initial assumptions, and unexpected outputs force pivots in research goals \crfix{(see Appendix~\ref{sec:related} for a detailed review of related work).}

To address these challenges, we introduce MatHarness, an agent harness spanning the full research lifecycle from initial ideation to paper writing. It rests on three design considerations. First, it ensures execution robustness through pre-execution verification with lightweight smoke runs, domain-expert skill abstractions, and just-in-time auditing hooks that intercept intermediate failures before errors compound. Second, it maintains epistemic adaptability by preserving research artifacts and maintaining an evolving scientific mental model, enabling agents to \crfix{iteratively revise their scientific beliefs}. Third, it leverages collective intelligence, deploying a multi-agent round table of heterogeneous foundation models for peer-reviewed ideation, with targeted human oversight at critical  checkpoints for scientific rigor and expert alignment. Collaborating with domain experts in computational catalysis, we evaluate MatHarness on adsorbate-induced Cu--Zn surface segregation (Section~\ref{sec:case-study}); the trajectory is analyzed transition by transition in Appendix~\ref{app:trajectory-details}, with claim-level evidence traceability in Appendix~\ref{app:claims}.

\section{System Design}
\label{sec:system}

Scaling agentic systems from bounded execution pipelines to open-ended, long-horizon scientific discovery introduces three systemic challenges. \textbf{First, long-horizon execution is fragile and prone to cascading error amplification.} Unflagged early misconfigurations compound exponentially downstream into massive computational waste, and under context-window limits an agent's actions can silently diverge from its research plan. \textbf{Second, unpredictable scientific trajectories require dynamic adaptability.} Non-convergence and unphysical anomalies demand real-time self-correction, while refuted hypotheses force experimental plans and even research objectives to evolve with the accumulating evidence. \textbf{Third, individual agents inevitably exhibit cognitive blind spots.} Single-agent architectures remain susceptible to domain-specific knowledge gaps and implicit biases that only external cross-validation exposes. Rather than a predefined workflow or an unconstrained execution loop, MatHarness pairs a dedicated mechanism \crfix{with} each challenge, decoupling robust execution from dynamic epistemic state management. \crfix{Implementation details for each mechanism are collected in Appendices~\ref{app:implementation}--\ref{app:human}, and the agent prompts in Appendix~\ref{app:prompts}.}

\begin{figure}[t]
    \centering
    \includegraphics[width=0.88\linewidth]{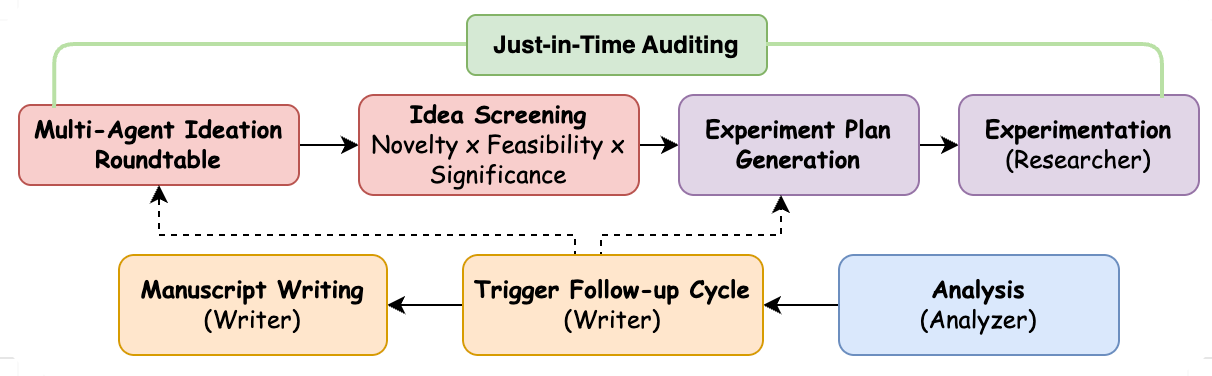}
    \vspace{-0.7em}
    \caption{The MatHarness ideation-to-paper cycle: a round table produces an executable
    experiment plan, the Researcher executes it under just-in-time auditing, and the writing
    agent synthesizes and audits the scientific story, triggering follow-up
    cycles (Appendix~\ref{app:implementation}).}
    \label{fig:pipeline}
    \vspace{-1.4em}
\end{figure}

\subsection{Ideation-to-Paper Research Cycle}

Given an open-ended scientific problem, MatHarness initiates a multi-agent round table to develop research proposals, screens them for novelty, feasibility and scientific significance, and refines the winning proposal into an executable experiment plan. The Researcher agent then executes it, while an independent Auditor agent monitors the trajectory for anomalies and drift; the Writer then synthesizes a paper plan while auditing the scientific story, pinpointing flawed hypotheses and surfacing new directions that trigger targeted follow-up cycles (Figure~\ref{fig:pipeline}). \rev{At each step within a cycle, the acting agent builds necessary context from the persistent state, proposes an action aligned with the active goal $G_t$, executes it, and updates its mental model. If an audit is triggered, the Auditor evaluates the trajectory to determine how to proceed next or whether to terminate}\crfix{.}

\subsection{Execution Robustness for Long-Horizon Trajectories}

\textbf{Planning before Execution.} Before dispatching jobs, MatHarness enforces a pre-execution planning phase that translates a scientific proposal into an explicit dependency graph of computational sub-tasks with per-job specifications. Instead of generating and blindly executing ad-hoc scripts, intended experiments undergo pre-flight verification: lightweight smoke runs plus prerequisite checks such as surface slab geometry, pseudopotential availability, and parameter consistency; this establishes checkpoints before full-scale runs, prevents premature resource allocation, and ensures runs adhere to the intended design.

\textbf{Skill Invocation for Rigorous Scientific Computing.} To ensure computational rigor and disciplined resource use, MatHarness equips agents with modular, domain-specific skills distilled from human experts. Rather than writing unconstrained code, agents invoke these skills to handle complex tasks (adsorbate site identification, MLIP surrogate evaluations, DFT calculations). Each skill encapsulates established conventions ($k$-point grid density, electronic convergence criteria) and best practices for compute infrastructure. \rev{Our case study (}\crfix{Section~\ref{sec:case-study}}\rev{) deployed skills covering materials modeling, HPC orchestration, fault diagnosis, and data analytics}\crfix{.} \rev{Furthermore, as new failure modes emerged during execution, human corrections were systematically codified into reusable skills.}

\textbf{Just-in-time Auditing.} Deferring auditing until an entire experiment completes is wasteful: an undetected early failure renders subsequent calculations invalid and forces an expensive restart. MatHarness instead places hooks at critical \rev{checkpoints} that trigger an independent Auditor, so defects are intercepted mid-execution and resolved by localized error recovery or plan adjustment before errors cascade. \rev{Specifically, the Auditor evaluates the research trajectory to prevent unintended goal drift, ensure experimental plans directly address the core research question, verify that execution strictly follows approved plans\crfix{,} and confirm that scientific claims are supported by empirical evidence. Guided by this evaluation, the Auditor issues targeted interventions, spanning continuation, validation, replanning, reframing, claim rejection, recovery, and escalation}\crfix{.}

\subsection{Epistemic Adaptability for Open-ended Discovery}

\textbf{Persistent Research Artifacts.} A research trajectory generates heterogeneous artifacts (hypotheses, execution logs, raw outputs, and failures) that are the raw material for downstream reasoning and verification. Preserving them establishes a traceable research lineage, giving agents observability into prior decision points and enabling strategic branching or hypothesis pivoting without redundant compute. To reconcile history with context limits, agents receive an index of these files and selectively retrieve only the context a task requires. \rev{Concretely, the research artifact $A_t$ captures the complete research state alongside raw execution outputs. It maintains active goals and plans, experiment outcomes, failure diagnoses, auditor findings, human interventions, and status-tagged scientific claims (tentative, validated, or rejected) linked to their supporting files}\crfix{.}

\textbf{Evolving Scientific Mental Model.} Unlike the artifact record, the scientific mental model holds the agent's cognitive state: evolving hypotheses, mechanistic interpretations, and unresolved uncertainties. As new computational evidence emerges, the agent dynamically updates or refutes the mental model, refining or pivoting high-level research strategies. \rev{We }\crfix{deliberately} \rev{maintain the mental model $M_t$ separately from the artifact $A_t$. While $A_t$ preserves objective execution history, $M_t$ represents the agent's current scientific interpretation, allowing the same evidence to be re-evaluated as new findings arrive. Rather than propagating $M_t$ sequentially across steps, the agent periodically reconstructs it from $A_t$. This decoupling ensures resilience against context truncation and prevents the persistence of beliefs no longer supported by evidence}\crfix{.}

\subsection{Collective Intelligence for Scientific Deliberation and Alignment}
\textbf{Multi-Agent Round Table.} To maximize cognitive diversity during ideation, MatHarness convenes a round table of agents powered by heterogeneous foundation models. Each agent independently synthesizes domain literature, formulates research directions, and critically assesses peer proposals through multi-turn debate, challenging mechanistic assumptions and highlighting feasibility risks. A \textit{Round-Table Chair} terminates discussion once consensus converges or an iteration threshold is reached. \rev{Three core rules are enforced to discipline multi-agent debate. First, every claim must cite its supporting basis or be explicitly flagged as unverified. Second, in the opening round, each agent must deliver a reasoned verdict on all peer proposals to ensure complete scrutiny. Third, in subsequent rounds, agents may speak only to contribute novel reasoning or evidence, preventing circular discussion and providing the Chair with a clear convergence criterion. Proposals that survive this deliberation undergo final screening for literature-backed novelty, technical feasibility, and scientific significance}\crfix{.}

\textbf{Human-Agent Synergy.} While MatHarness emphasizes execution autonomy, the capability bounds of current foundation models and the need for interpretability necessitate targeted human oversight. MatHarness activates a human-in-the-loop mechanism at critical checkpoints, such as approving candidate proposals, validating unpredicted physical anomalies, or confirming strategic hypothesis pivots. Structured summaries at these junctures minimize expert burden while keeping exploration interpretable, safe, and aligned. \rev{Humans intervene through the persistent state by injecting an authoritative guidance message or by approving or rejecting an agent proposal at a decision point. }\crfix{Because humans may neither edit files directly nor steer the agent off the record, every intervention is logged in $A_t$, so all later conclusions remain traceable to agent reasoning or human decisions.}

\section{Case Study}
\label{sec:case-study}

\textbf{Research Problem Background.} Copper-zinc (Cu–Zn) alloys serve as the benchmark catalyst for industrial methanol synthesis, yet the chemical state and spatial distribution of Zn under reaction conditions remain a fundamental open question~\citep{behrens2012, kuld2016, kattel2017, amann2022}. The catalytic interface continuously restructures in response to gas-phase reactants and intermediates~\citep{tao2008}. Understanding these adsorbate-driven mechanisms is essential for rational catalyst design, since surface composition dictates active-site structure under operating conditions.

\textbf{This problem is a useful test bed for an agentic system}: navigating the high-dimensional combinatorial space of adsorbates and reaction configurations is bottlenecked by cognitive bias and tedious calculation, while the investigation itself requires iteratively refining hypotheses, deploying non-trivial computational methods, and deriving insight from open-ended exploration. Full computational protocol and execution budget in Appendix~\ref{app:case-details}.

\rev{\textbf{Data Provenance.} All results presented in this section derive exclusively from experiments planned, executed, and audited by MatHarness, incorporating no external data from human collaborators or the literature. We present the generated manuscript verbatim in Appendix \ref{app:generated-paper}, with its core scientific claims verified by domain experts. }\crfix{The prose is left unedited to show MatHarness's authentic output.} \rev{For instance, the manuscript's reference to Cu-rich data as a }\crfix{``companion study''} \rev{stems from a mid-run goal revision that temporarily split the composition limits into separate drafts before re-merging them; the phrase thus denotes MatHarness's own prior draft rather than external work.}

\textbf{MatHarness Trajectories and Research Outcomes.} MatHarness began with pre-execution convergence testing of electronic parameters and slab geometries, establishing a foundation for MLIP-prescreened production DFT. At the Cu-rich limit, clean surfaces favor Zn segregation, but automated auditing filtered structural reconstructions and revealed that most intermediates (14 of 17 science-set points) suppress surface Zn (most strongly HCO and O on Cu(211)), governed by the differential-adsorption identity $\Delta\Delta E_{\text{seg}} = E_{\text{ads}}^{\text{Zn}} - E_{\text{ads}}^{\text{Cu}}$ (minority Zn; the labels exchange in the reciprocal Zn-rich limit). A semi-quantitative thermodynamic model estimated how the segregation tendency varies across temperature and adsorbate-pressure conditions. Expanding to the Zn-rich limit, across the sampled Zn-rich facets  MatHarness found that $\text{CO}$ reverses the clean-surface preference for bulk-like Cu placement, with $d$-band shifts and Bader-charge trends consistent with stronger $\text{Cu--CO}$ hybridization. Unifying both limits, MatHarness concluded that adsorbates drive both composition endpoints toward a shared tendency of greater  surface Cu exposure within the sampled dilute-limit models and compiled the findings into a complete manuscript (Appendix~\ref{app:generated-paper}; claim-level traceability in Appendix~\ref{app:claims}).

\textbf{Trajectory Analysis.}
We evaluate the trajectory across completion, failure recovery, and evidence traceability \crfix{(Table~\ref{tab:main-summary}; DFT cell-runs include convergence tests and restarts).}

\begin{crblock}
\begin{table}[t]
    \centering
    \footnotesize
    \setlength{\tabcolsep}{4pt}
    \caption{Trajectory statistics (left) and all human interventions (right). ``$\to$ skill'': codified into a reusable skill on first occurrence; never recurred (Appendices~\ref{app:human}, \ref{app:case-details}).}
    \label{tab:main-summary}
    \begin{tabular}{lr@{\hspace{14pt}}lr}
        \toprule
        \textbf{Trajectory} & & \textbf{Human intervention} & \textbf{\#} \\
        \midrule
        Tool calls & 13{,}679 & Initial specification & 1 \\
        Plan / goal revisions & 6 / 2 & Local correction ($\to$ skill) & 3 \\
        DFT cell-runs / GPU-hours & 191 / $\sim$970 & Method constraint ($\to$ skill) & 1 \\
        MLIP prescreen evaluations & 171 & Goal steering & 2 \\
        Production DFT (force / plateau) & 50 (21 / 29) & Fork approval / escalation & 9 \\
        Rejected by reconstruction gate & 8 & \textbf{Total} & \textbf{16} \\
        \bottomrule
    \end{tabular}
    \vspace{-0.6em}
\end{table}
\end{crblock}

\textbf{Human Oversight.} \crfix{Table~\ref{tab:main-summary} reports all 16 human interventions, including nine decision-fork approvals (e.g., approving a validation sub-campaign, redirecting compute to the formate coverage series). Each technical fix (hardware routing, process-kill guardrails, the reconstruction gate, the resume fix) was codified into a skill on first occurrence and never recurred, so ongoing human effort concentrated on strategic governance.}

\textbf{\crfix{Self-Correction in Practice.}} Key discovery branches emerged from intermediate evidence rather than a pre-encoded plan. \crfix{\emph{(i) Hypothesis revision.} The MLIP prescreen of 171 configurations suggested that most intermediates leave Zn segregation largely unchanged; converged DFT reversed this picture (14 suppress, 2 promote, 1 neutral on the 17-point science set), so the agent revised $M_t$ and demoted the surrogate to a queue-ordering tool. \emph{(ii) Parameter consistency.} The Auditor blocked a facet ranking computed with per-facet $k$-point meshes and enforced a uniform per-cell density ($k\times\mathrm{cell}\approx 50$--$51$), which became a standing convergence gate. \emph{(iii) Claim correction.} The draft's centerpiece claim was that facet geometry flips the sign of the formate effect ($-0.43$~eV on Cu(100) vs.\ $+0.14$~eV on Cu(211)). The Auditor noted that the two points also differ in coverage ($\theta=0.19$ vs.\ $0.08$) and triggered a matched-coverage series: formate promotes on Cu(100) at every coverage tested and on Cu(211) when dilute ($-0.12$~eV at $\theta\approx0.04$), reversing only at higher coverage. The facet attribution was overturned before publication; facet geometry instead sets the coverage at which the sign reverses. \emph{(iv) Failure recovery.} A relaxation resumed from a truncated checkpoint looped silently on a GPU; MatHarness traced it to a corrupted optimizer history, restarted from the intact density checkpoint, and codified a permanent resume rule (Appendix~\ref{app:trajectory-details}).}

\textbf{\crfix{Controlled Comparison.}} On a small controlled materials-audit battery, MatHarness accepted more valid cases and caught more seeded flaws than a single-turn standalone LLM baseline \crfix{(valid cases passed: 4/4 vs.\ 0/4; seeded flaws caught: 4/4 vs.\ 2/4; no false rejections),} while mental model augmentation increased true-flaw recall from 10/16 to 15/16 (Appendix~\ref{app:evaluation}). Domain collaborators estimate a comparable campaign would occupy an experienced PhD researcher for roughly one to two month\rev{s} while MatHarness takes about two weeks.

\textbf{\crfix{Limitations} and Discussion.} Several physical and systemic boundaries define the scope of our conclusions. From a materials science perspective, the derived insights remain conditional on the fidelity of standard exchange-correlation functionals (e.g., PBE), idealized metallic dilute-limit models, and the numerical quality of the underlying relaxations: 21 of 50 production relaxations met the strict force criterion and 29 reached a stable energy plateau under the capped protocol, the latter retained as best-effort calculations rather than force-converged structures (Appendix~\ref{app:case-details}). Headline trends are therefore restricted to the auditor-vetted science set and are qualitative-to-semi-quantitative. From a system design perspective, \crfix{two evaluation gaps remain. First, the only baseline is a tool-less single-turn LLM on an eight-case audit battery: we did not compare against prior agent systems, as rerunning them over a multi-week DFT campaign exceeds our compute budget, and per-mechanism ablations remain future work. Second, the evidence rests on a single trajectory, so run-to-run variance arising from LLM stochasticity remains unquantified.} We therefore position this work not as a claim that autonomous materials discovery is solved, but as evidence from a single end-to-end case study that a structured agentic harness can sustain and adapt a non-linear, hypothesis-led computational research trajectory.

\newpage
\bibliographystyle{plainnat}
\bibliography{references}

\clearpage
\appendix

\section*{Appendix}

\section{Appendix Overview}
\label{app:overview}

The supplementary material is organized into four groups.

\paragraph{Positioning and system design.}
Appendix~\ref{sec:related} reviews related work, identifying the gaps between existing frameworks and domain-specific needs that motivate MatHarness. 
Appendix~\ref{app:implementation} describes the MatHarness implementation and execution lifecycle, including the end-to-end pipeline of Figure~\ref{fig:pipeline} and the execution environment.
Appendix~\ref{app:skills} details the domain-specific skills exposed to the research agent. 
Appendix~\ref{app:memory} describes the persistent research artifacts and the evolving scientific mental model. Appendix~\ref{app:auditor} describes just-in-time auditing, the auditor's
consistency criteria and possible outcomes.
Appendix~\ref{app:human} summarizes human--agent interaction and separates
one-time capability building from recurring per-run oversight.

\paragraph{Case study.}
Appendix~\ref{app:case-details} provides the complete alloy-segregation
experimental protocol, computational model, execution budget, and trajectory statistics. Appendix~\ref{app:trajectory-details} analyzes representative research transitions and failures, including the auditor-triggered correction that overturned an over-attributed claim.  Appendix~\ref{app:claims} traces each final scientific claim to its supporting computation and validation route.

\paragraph{Provenance and evaluation.}
Appendix~\ref{app:reproducibility} documents provenance \rev{(including the data-provenance statement summarized in Table~\ref{tab:provenance})}, software and model
versions, third-party assets and licenses, and execution cost.
Appendix~\ref{app:evaluation} reports trajectory-level metrics, the controlled
comparison against a standalone LLM, and the ablations that remain future work.

\paragraph{Generated artifacts.}
Appendix~\ref{app:prompts} reproduces the agent instructions.
Appendices~\ref{app:state-snapshot}, \ref{app:auditor-example},
and~\ref{app:artifacts} reproduce, respectively, a research-state snapshot, an
auditor record, and the artifact manifest taken from the run.
Finally, Appendix~\ref{app:generated-paper} contains the scientific manuscript produced by MatHarness during the case study.

\newpage
\section{Related Work}
\label{sec:related}

\paragraph{Domain-specific Scientific Agents for Material Science.} AtomAgents \citep{ghafarollahi2025atomagents} integrates retrieval, multimodal analysis, and atomistic simulation for alloy design; MatSciAgent \citep{chaudhari2026matsciagent} coordinates agents for retrieval, structure generation, continuum simulation, and molecular dynamics; and \citet{vriza2026atomisticsim} automate atomistic workflows on HPC resources. MAPPS \citep{zhou2025mapps} combines LLM planning, physics-grounded code generation, and human-mediated recovery, while HoneyComb \citep{zhang2024honeycomb} couples a materials knowledge base with computational
tools. Adsorb-Agent \citep{ock2026adsorbagent} and LLMatDesign \citep{jia2024llmatdesign} target specific tasks of stable adsorption-configuration
search and iterative target-property materials design, respectively.

These systems primarily automate \emph{task execution}: select tools, run calculations, and return results given a defined task. MatHarness instead targets a much broader spectrum of the research lifecycle, from scientific problem formulation and research planning to iterative experiment design, execution, validation, interpretation, and writing.

\paragraph{General-purpose AutoResearch systems.}
Building on the reason-and-act loop underlying most tool-using agents \citep{yao2023react}, AI Scientist \citep{lu2024aiscientist} and AI
Scientist-v2 \citep{yamada2025aiscientistv2} automate idea generation, experimentation, paper writing, and review. Karpathy's AutoResearch \citep{karpathy2026autoresearch} iteratively modifies and evaluates code changes within an experimental loop. Agent Laboratory \citep{schmidgall2025agentlab} and EvoScientist \citep{lyu2026evoscientist} coordinate multiple agents across research stages. Co-Scientist \citep{gottweis2025coscientist} focuses on hypothesis generation and ranking through multi-agent debate. Robin \citep{ghareeb2026robin} integrates literature search, experiment design, and data analysis for biomedical discovery. Latest systems address long-horizon persistence: AutoSci \citep{qian2026autosci} combines structured research memory with a staged lifecycle and evolving research skills, while ScienceFlow \citep{zhao2026scienceflow} represents research as recoverable executable states supporting branching, backtracking, and resource-aware execution.

While these general-purpose systems demonstrate the feasibility of end-to-end research, applying long-horizon AutoResearch to computational materials science introduces distinct challenges. First, agents must navigate complex, domain-specific computational protocols and simulation settings that leave little room for error. Second, sustained long-horizon execution demands execution robustness and logical consistency to prevent drift and cascading failures~\citep{fei2026agents}. Third, agents must dynamically adapt their research goals based on intermediate results, which requires maintaining fine-grained control over both an execution state and an epistemic state. MatHarness addresses these challenges through a stateful harness with  evidence-driven goal evolution, just-in-time auditing, and long-horizon memory management.

\paragraph{Automated validation and falsification.}
A parallel line of work automates the \emph{checking} of scientific claims rather than their generation. Popper \citep{huang2025popper} validates free-form hypotheses through agentic sequential falsification with statistical error control, and AIGS \citep{liu2024aigs} makes falsification an explicit role inside the discovery loop. MatHarness shares the premise that generation without adversarial checking is unsafe, but differs in where the check draws its independence: rather than re-examining a claim with the same model or a statistical test over textual evidence, the MatHarness auditor re-derives load-bearing results with an independent computational engine---a second interatomic potential, and where feasible a DFT cross-check---so that agreement carries physical rather than rhetorical weight (Appendix~\ref{app:auditor}).

\newpage
\section{MatHarness Implementation Details}
\label{app:implementation}

\subsection{Research lifecycle}

A MatHarness run begins from a research specification containing the initial
scientific objective \(G_0\), user-provided constraints, an execution workspace,
and the available tools and domain-specific skills. A complete research trajectory may contain multiple research iterations. Each iteration
contains the following conceptual stages:

\begin{enumerate}
    \item \textbf{Context construction.}
    The harness selects task-relevant information from the persistent research
    state and workspace.

    \item \textbf{Scientific reasoning.} Based on existing research artifacts and mental model, the agents plan the next scientifically meaningful action.

    \item \textbf{Planning check.}
    The proposed action is evaluated relative to the active research goal and
    current plan.

    \item \textbf{Action execution.} Agents may invoke tools and skills, and execute generated scripts.

    \item \textbf{Observation ingestion.}
    Generated outputs, errors, analyses, and metadata are incorporated into the
    persistent research artifacts.

    \item \textbf{Scientific interpretation.}
    The research agent updates its hypotheses, uncertainty, and interpretation
    of the accumulated evidence.

    \item \textbf{Validation and auditing.}
    Relevant mechanical checks and just-in-time-auditor reviews are performed.

    \item \textbf{Continuation, replanning, or escalation.}
    The system either continues the existing plan, modifies the active goal or
    plan, requests human input, or terminates the research run.
\end{enumerate}

This lifecycle is intentionally not a rigid scientific workflow. Individual
iterations can differ substantially in duration and computational complexity,
and multiple low-level computational operations may occur within a single
research step.

\subsection{End-to-end ideation-to-paper pipeline}
\label{app:pipeline}

The end-to-end Ideation-to-Paper Pipeline, as implemented in the MatHarness research cycle, aims to accelerate scientific discovery from an open-ended research problem to a manuscript with scientifically meaningful contributions. Crucially, while the pipeline encompasses structured phases, it fundamentally departs from conventional predefined or static workflows. Instead of relying on hardcoded step transitions, our framework operates as a dynamic, event-driven, closed-loop architecture grounded in execution robustness, epistemic adaptability, and collective intelligence. Oversight across the entire lifecycle is maintained by a global just-in-time (JIT) auditing mechanism that continuously evaluates research progress, actively reshaping the execution trajectory in real time.

Given an open-ended research problem, specialized LLM agents conduct multi-agent round-table discussions under the moderation of a Round-Table Chair to refine raw concepts into precise, testable hypotheses. Following ideation, abstract research goals are translated into executable workflows governed by explicit dependency graphs. Lightweight smoke runs are executed prior to deployment to validate environment configurations, code logic, and computational tool access.

The Researcher agent then programmatically invokes specialized computational tools, such as Density Functional Theory (DFT) solvers \rev{and} Machine Learning Interatomic Potentials (MLIP), and \rev{generates} experimentation scripts. The Auditor agent tracks execution progress and performs local error recovery, dynamically handling runtime exceptions. As execution proceeds,  raw outputs are preserved as persistent research artifacts, which are cataloged in an artifact index, creating a traceable research lineage that connects source data directly to generated figures and empirical claims. The Analyzer agent analyzes empirical results and organizes raw outputs into structured findings and insights. Throughout the process, the mental model is updated, continuously testing empirical results against active hypotheses to refine mechanistic interpretations and domain understanding.

\rev{The Writer agent actively eliminates ungrounded claims, verifies logical continuity, and ensures citation accuracy.} If scientific uncertainties, anomalous data points, or promising new research directions are detected, the Writer agent initiates follow-up research cycles, routing new questions back into the multi-agent deliberation stage for iterative expansion and further experimentation.

Human-agent synergy is integrated through critical human-in-the-loop checkpoints throughout the entire process, allowing domain experts to evaluate proposals, identify potential anomalies, and authorize strategic direction pivots.

\subsection{Execution environment}

The computational environment used in our experiments contains the following
capabilities:

\begin{itemize}
    \item \textbf{Underlying frontier large language model:} the LLMs used for different research stages are summarized in Table\rev{~}\ref{tab:software}.

    \item \textbf{Atomistic simulation:}
    plane-wave DFT with Quantum ESPRESSO~\citep{qe2017} (GPU builds); ASE~\citep{ase2017} for surface, adsorbate, and
    slab construction; machine-learning interatomic potentials (NequIP~\citep{nequip2022} for pre-screening,
    MACE~\citep{mace2022} and CHGNet~\citep{chgnet2023} for independent cross-checks) for cheap configuration surveys.

    \item \textbf{Interatomic or electronic model:}
    PBE exchange--correlation~\citep{pbe1996} with GBRV ultrasoft pseudopotentials~\citep{gbrv2014} and a DFT-optimised Cu
    lattice constant ($a_0 = 3.6366$~\AA); NequIP/MACE/CHGNet for surrogate energetics.

    \item \textbf{Scientific analysis:}
    a Python stack---NumPy, pandas, Matplotlib, ASE, scikit-learn (descriptor regression
    with leave-one-adsorbate-out cross-validation)---plus custom semi-grand-canonical
    Monte Carlo and cluster-expansion fitting.

    \item \textbf{Literature access:}
    web and literature retrieval used for novelty pre-screening and to anchor the metric
    ladder and validation discipline against prior work.

    \item \textbf{Persistent workspace:}
    a shared cluster filesystem plus a git repository and structured JSON records
    (per-cell \texttt{result.json}, an aggregated \texttt{analysis\_master.json}, an
    \texttt{experiment\_plan.json}, and audit records), together with a durable
    file-based memory store carried across model contexts.

    \item \textbf{Resource limits:}
    a heterogeneous GPU pool (H100 and V100 for DFT; a policy explicitly forbidding a
    slower GPU class for DFT); per-cell wall-time and ionic-step caps with graceful
    checkpoint/stop; and a monthly token budget on the auditor LLM endpoint.
\end{itemize}

All computational artifacts required to reconstruct the final analysis are
retained in the run workspace under a no-delete-without-backup policy: raw outputs
(\texttt{espresso.pwo}), per-cell results, and analysis artifacts are never overwritten
or deleted without an explicit backup, while large scratch wavefunctions are written to
node-local disk and may be recycled.

\newpage
\section{Domain-Specific Skills}
\label{app:skills}

Generic tool use provides execution capability but does not necessarily encode
how computational materials scientists combine those capabilities into robust
scientific procedures. MatHarness therefore includes a collection of
domain-specific skills.

Each skill is defined by a structured specification containing, where
applicable:

\begin{itemize}
    \item the scientific purpose of the skill;
    \item expected inputs and preconditions;
    \item the low-level tools or software it may invoke;
    \item recommended computational procedure;
    \item expected outputs and generated artifacts;
    \item domain-specific validation checks;
    \item common failure modes;
    \item interpretation guidance; and
    \item conditions under which human escalation is recommended.
\end{itemize}

Table~\ref{tab:skills} lists the domain-specific skills used in the
Cu-Zn alloy-segregation study.

\begin{table}[ht]
    \centering
    \small
    \caption{Domain-specific skills used in the Cu-Zn alloy-segregation study.}
    \label{tab:skills}
    \begin{tabular}{p{0.27\linewidth}p{0.62\linewidth}}
        \toprule
        \textbf{Skill} & \textbf{Scientific role} \\
        \midrule
        Structure construction &
        Construct and validate surfaces, interfaces, grain boundaries, or
        alloy configurations under explicit geometric constraints. \\

        Convergence analysis &
        Test sensitivity to supercell size, sampling, relaxation criteria,
        trajectory length, or related numerical parameters. \\

        Segregation analysis &
        Compute segregation-related energetic or compositional observables
        using consistent reference states and normalization. \\

        Simulation diagnosis &
        Inspect simulation logs, trajectories, convergence behavior, and
        numerical instabilities to diagnose failed calculations. \\

        Statistical validation &
        Estimate uncertainty, compare repeated calculations, and determine
        whether observed trends exceed numerical or statistical noise. \\

        Scientific visualization &
        Produce plots or structural visualizations tied to explicit scientific
        questions and stored provenance. \\

        Compute infrastructure orchestration & Manage high-performance computing (HPC) environments, submit parallel batch jobs, and dynamically allocate compute resources across heterogeneous clusters. \\
        \bottomrule
    \end{tabular}
\end{table}

Skills are not intended to replace scientific reasoning. The research agent
still decides whether a skill is appropriate, how its outputs should be
interpreted, and whether additional controls are needed. Skills instead reduce
the amount of procedural knowledge that must be reconstructed in each model
context and provide a consistent interface to recurring scientific practices.

\newpage
\section{Long-Horizon Memory}
\label{app:memory}

\subsection{Persistent Research Artifacts}

The persistent research artifact acts as the durable project record. \rev{The artifact $A_t$ at time $t$ contains:}

\begin{itemize}
    \item the initial research objective \(G_0\);
    \item the current active research goal \(G_t\);
    \item the current plan and milestone status;
    \item completed, active, and proposed experiments;
    \item simulation and analysis artifacts;
    \item observations and validated evidence;
    \item failed experiments and failure diagnoses;
    \item open scientific questions;
    \item accepted, tentative, and rejected scientific claims;
    \item unresolved uncertainties;
    \item human interventions;
    \item auditor findings; and
    \item provenance links between decisions and evidence.
\end{itemize}

The exact implementation \rev{distributes} this information across structured
records and the persistent workspace rather than storing it in one serialized
object.

\subsection{Context Reconstruction On Demand}

A central design goal is that the full research trajectory need not be replayed into the model context at every iteration. Instead, the harness reconstructs task-relevant context from persistent state.

The reconstruction policy considers:

\begin{itemize}
    \item the current goal and active plan;
    \item evidence relevant to the current decision;
    \item unresolved contradictions;
    \item recent experiments;
    \item historical failures relevant to a proposed action;
    \item current scientific claims and their evidential status; and
    \item references to large artifacts that can be inspected on demand.
\end{itemize}

This allows the project to survive context truncation, model restarts, or
changes in the underlying agent configuration without losing the durable state
of the research process.

\subsection{Mental world model}

The mental world model \(M_t\) complements the persistent research state by
representing the agent's current scientific belief and domain understanding. It may include:

\begin{itemize}
    \item active and competing hypotheses;
    \item inferred physical mechanisms;
    \item causal or explanatory relationships between observations;
    \item confidence in current conclusions;
    \item unresolved ambiguity;
    \item expected outcomes of candidate experiments; and
    \item the agent's current explanation of why the active goal is worth
    pursuing.
\end{itemize}

The distinction between \(A_t\) and \(M_t\) is important. For example, two
agents may have access to the same experiment history \(A_t\) but interpret the
evidence differently, yielding different \(M_t\) and therefore different next
experiments. Conversely, a newly completed experiment may update the persistent
record without substantially changing the scientific interpretation.

For this reason, MatHarness periodically reconstructs or revises the mental
world model from the persistent evidence rather than treating previous model
reasoning as authoritative.

\newpage
\section{Long-Horizon Robustness and Just-in-Time Auditing}
\label{app:auditor}

\subsection{Auditor role}

The live auditor provides an independent assessment of the current research
trajectory. The auditor receives a structured view of the research objective,
active goal, plan, recent actions, relevant evidence, scientific claims, and
unresolved issues.

The auditor evaluates \crfix{five} categories of consistency:

\paragraph{Goal consistency.}
Does the active research goal remain scientifically connected to the initial
objective? If the goal has changed, is the change supported by identifiable
evidence or an explicit human instruction?

\paragraph{Plan consistency.}
Do proposed actions contribute to resolving an active scientific question, or
has the trajectory accumulated locally interesting but strategically irrelevant
work?

\paragraph{Execution consistency.}
Do actual computational runs, code edits, and generated deliverables faithfully align with the approved plan? Has the agent suffered execution drift, taken unauthorized shortcuts, or silently dropped specified tasks and evaluation criteria?

\paragraph{Evidence consistency.}
Are scientific claims supported by reproducible calculations and appropriate
validation? Have contradictory observations been ignored or prematurely
dismissed?

\paragraph{Epistemic consistency.}
Are hypotheses, observations, and conclusions kept distinct? Has speculative
reasoning been promoted into accepted scientific state without adequate
support?

\subsection{Possible auditor outcomes}

An audit can result in one of the following actions:

\begin{enumerate}
    \item \textbf{continue}: no material issue is detected;
    \item \textbf{validate}: request a missing convergence test or control;
    \item \textbf{replan}: revise the research plan while preserving the active
    goal;
    \item \textbf{reframe}: reconsider the active research goal;
    \item \textbf{reject}: remove or downgrade an unsupported scientific claim;
    \item \textbf{recover}: revisit an earlier stable research state after a
    problematic branch; or
    \item \textbf{escalate}: request human judgment.
\end{enumerate}

\subsection{Auditing frequency}

The auditor is invoked under a hybrid policy: event-triggered by the situations below, and additionally as a standing periodic pass that independently recomputes the key energetic results with a different computational engine.
Possible trigger events include:

\begin{itemize}
    \item completion of a major research milestone;
    \item proposal of a significant goal change;
    \item repeated execution failures;
    \item emergence of contradictory evidence;
    \item promotion of a new scientific claim;
    \item substantial computational expenditure;
    \item attempted research termination; and
    \item explicit human request.
\end{itemize}

In the Cu-Zn alloy-segregation study, the most
consequential actions of the auditor are: enforcing a single per-cell $k$-point density across facets
after detecting an inconsistent-parameter comparison; rejecting a machine-learning-potential
conclusion once the converged-$k$ DFT contradicted it; \rev{applying the human-specified physical discernment
gate ($|\Delta\Delta E_{\mathrm{seg}}| > 0.5$~eV; a method constraint in Appendix~\ref{app:human})} that removed eight reconstruction
blow-ups from the science set; and flagging a facet-versus-coverage confound in the
formate result, triggering a targeted coverage series that ultimately \emph{overturned} the
facet sign-flip reading. The consistency-and-convergence check is implemented as a standing
tool, \texttt{audit\_convergence.py}, run as a CI-style gate over the campaign: on a 42-point
audit it confirmed a uniform per-cell $k$-point density and auto-flagged every under-converged
$\Delta E_{\mathrm{seg}}$, exiting with non-zero status so that an inconsistent or unconverged
comparison cannot pass silently. Independently of these events, the
auditor re-derived the sign of the key $\Delta\Delta E_{\mathrm{seg}}$ trend with a
second interatomic potential (CHGNet against MACE) to expose model dependence.

\newpage
\section{Human--Agent Interaction}
\label{app:human}

Human interaction is treated as a first-class event in the research trace.
Interventions are categorized by their role rather than being reported only as
a total count.

\begin{table}[ht]
    \centering
    \small
    \caption{Human-intervention categories.}
    \label{tab:human-categories}
    \begin{tabular}{p{0.25\linewidth}p{0.63\linewidth}}
        \toprule
        \textbf{Category} & \textbf{Description} \\
        \midrule
        Initial specification &
        Research question, scientific scope, constraints, or available
        computational resources. \\

        Domain clarification &
        Missing scientific information or interpretation supplied by a human. \\

        Local correction &
        Correction to a simulation setup, software configuration, or analysis. \\

        Goal steering &
        Reprioritization, refinement, or redefinition of the active research
        goal. \\

        Method constraint &
        Human specifies that a method, model, or computational route should or
        should not be used. \\

        Auditor escalation &
        Human resolves an uncertainty or conflict that the system cannot
        justify autonomously. \\

        Termination decision &
        Human stops or redirects the research process. \\
        \bottomrule
    \end{tabular}
\end{table}

\begin{table}[!ht]
    \centering
    \small
    \caption{\rev{Human interventions in the case study.}}
    \label{tab:human-ledger}
    \begin{tabular}{p{0.24\linewidth}cp{0.50\linewidth}}
        \toprule
        \textbf{Category} & \textbf{\#} & \textbf{Description} \\
        \midrule
        Initial specification & 1 &
        The scientific objective (quoted in Appendix~\ref{app:case-details}). \\

        Local correction \newline ($\to$ skill) & 3 &
        Hardware routing for DFT by system size; process-kill guardrails
        (PID / working-directory-strict, no-kill on shared GPUs); the checkpoint
        resume-death-loop fix. \\

        Method constraint \newline ($\to$ skill) & 1 &
        The $|\Delta\Delta E_{\mathrm{seg}}| > 0.5$~eV reconstruction validity gate. \\

        Goal steering & 2 &
        Thesis elevation (intermediate-only $\to$ intermediate-and-facet) and the
        two-paper split; matches the explicit goal revisions in
        Table~\ref{tab:trajectory-detailed}. \\

       \rev{Auditor escalation} \newline \rev{/ Fork approval} & 9 &
       \rev{Approvals at decision forks}\crfix{, itemized below by stage (F1--F9).} \\
      
       & & \crfix{F1 (Scope):} \tbv{select the round-table proposal to pursue and sign off on the executable experiment plan.} \\
       & & \crfix{F2 (Convergence):} \tbv{adopt the Auditor's escalated uniform per-cell $k$-point density as a standing gate.} \\
       & & \crfix{F3 (MLIP pre-screen):} \tbv{allocate the production-DFT compute budget to the surrogate-ranked queue.} \\
       & & \crfix{F4 (Calibration):} \tbv{approve the MLIP-vs-DFT validation sub-campaign.} \\
       & & \crfix{F5 (Calibration):} \tbv{confirm demoting the MLIP surrogate to a queue-ordering tool after DFT reversed its prediction.} \\
       & & \crfix{F6 (Production DFT):} \tbv{accept energy-plateau relaxations as best-effort results under the capped protocol.} \\
       & & \crfix{F7 (Production DFT):} \tbv{authorize extending the study from the Cu-rich to the Zn-rich composition limit.} \\
       & & \crfix{F8 (Analysis):} \tbv{redirect compute from a low-priority reaction-barrier calculation to the matched-coverage formate series.} \\
       & & \crfix{F9 (Manuscript):} \tbv{sign off on the post-audit claim set (facet attribution withdrawn; phase-diagram claims downgraded to qualitative).} \\
        \midrule
       \crfix{\textbf{Total}} & \crfix{16} & \\
        \bottomrule
    \end{tabular}
\end{table}

A human intervention updates the same persistent state used by autonomous
research. Consequently, the system can continue from the modified state rather
than restarting the research process. Human input also becomes part of the
provenance record so that later conclusions can be traced to both autonomous
and human decisions.

\subsection{Human intervention in the alloy-segregation case study}

Human interventions in this study fall into two qualitatively distinct categories; conflating them artificially inflates the perceived steady-state operational burden. To accurately reflect continuous agent autonomy, we decompose total intervention $H$ into one-time capability building and per-run oversight:
\begin{equation}
H = H_{\mathrm{cap}} + H_{\mathrm{run}}
\end{equation}

One-time capability-building interventions ($H_{\mathrm{cap}}$) amortize toward zero over subsequent campaigns. These encompass technical corrections that were codified into reusable skills or execution runbooks upon their initial occurrence—such as system-size hardware routing for DFT, process-kill safety guardrails (e.g., directory-strict execution and protecting shared GPUs), a $|\Delta\Delta E_{\mathrm{seg}}| > 0.5$~eV structural reconstruction gate, and checkpoint resume-loop fixes. Mapping primarily to the \emph{local correction} and \emph{method constraint} categories in Table~\ref{tab:human-categories}, these technical fixes represent a fixed, front-loaded setup cost that does not recur once encoded.

Conversely, per-run irreducible interventions ($H_{\mathrm{run}}$) capture the true ongoing human overhead during steady-state operation. $H_{\mathrm{run}}$ comprises the initial research objective specification and a minimal set of high-level \emph{goal-steering} and \emph{termination} gates at critical strategic forks. Examples include approving validation sub-campaigns, reallocating compute from low-priority reaction-barrier calculations toward a targeted formate coverage series, and elevating the thesis from an intermediate-only scope to a joint intermediate-and-facet account. Thus, $H_{\mathrm{run}}$ serves as the authentic metric for continuous agent autonomy, distinguishing essential strategic oversight from routine technical maintenance.

\rev{Table~\ref{tab:human-ledger} itemizes human interventions in the case study.}

\clearpage

\section{Alloy-Segregation Case Study: Experimental Details}
\label{app:case-details}

\subsection{Scientific objective}

The initial research objective is:

\begin{quote}
Investigate how the reaction intermediates of CO$_2$ hydrogenation reshape the surface
segregation of a Zn dopant in Cu, across the low- and high-index facets a working
nanoparticle exposes and across a realistic coverage range; determine which intermediates
promote or suppress Zn surface segregation, establish the governing mechanism, and place
the result in working temperature--pressure conditions, producing a publishable analysis.
\end{quote}

We intentionally provide the system with a research objective rather than a
step-by-step computational protocol.

The investigated system is a dilute Zn dopant in Cu, modelled as periodic slabs of the
(111), (100), (110), and (211) facets, with ten CO$_2$-hydrogenation intermediates
(O, OH, H$_2$O, H, CO, CO$_2$, HCO, HCOO, H$_2$COO, CH$_3$OH) spanning the accepted
CO$_2$-to-methanol pathway on Cu~\citep{grabow2011}, over a coverage ladder, evaluated
in a working regime of $100$--$700\,^\circ$C and $10^2$--$10^6$~Pa. The primary
scientific quantities are the clean segregation energy $E_{\mathrm{seg}}$, the
adsorbate-induced shift $\Delta\Delta E_{\mathrm{seg}}$, the environmental segregation
energy $E_{\mathrm{eseg}}(T,p)$, evaluated within the \emph{ab initio}
atomistic-thermodynamics framework~\citep{reuter2001}, and the equilibrium Zn coverage from semi-grand-canonical
Monte Carlo. The final conclusion is assessed against an independent interatomic potential
(CHGNet vs MACE), a physical discernment gate for reconstructions, leave-one-adsorbate-out
cross-validation of descriptor models, and consistency with prior water-on-transition-metal
segregation trends in the literature~\citep{ruban1999,ye2024}.

\subsection{Computational model}

The case study uses:

\begin{itemize}
    \item \textbf{Structure representation:} periodic Cu--Zn slabs with a single Zn
    placed either at the surface (segregated) or in the bulk (substitutional), plus
    adsorbed intermediates at controlled coverage.
    \item \textbf{Simulator:} plane-wave DFT (Quantum ESPRESSO, GPU).
    \item \textbf{Potential / electronic-structure model:} PBE with GBRV ultrasoft
    pseudopotentials; NequIP/MACE/CHGNet surrogates for pre-screening and cross-checks.
    \item \textbf{Boundary conditions:} periodic slabs with a vacuum gap and fixed
    bottom layers; a uniform per-cell $k$-point density ($k\times\mathrm{cell}\approx
    50$--$51$) on Monkhorst--Pack grids~\citep{monkhorst1976} enforced across facets for direct comparability.
    \item \textbf{Relaxation or sampling protocol:} damped BFGS geometry relaxation with
    a convergence gate distinguishing true convergence from energy-plateau best-effort;
    semi-grand-canonical Monte Carlo for equilibrium coverage.
    \item \textbf{Temperature / composition conditions:} $100$--$700\,^\circ$C,
    $10^2$--$10^6$~Pa, coverage $\theta \approx 0.04$--$0.375$.
    \item \textbf{Analysis quantities:} $\Delta\Delta E_{\mathrm{seg}}$ per (facet,
    intermediate, coverage), $E_{\mathrm{eseg}}(T,p)$, SGCMC coverage, and
    descriptor-model generalization under leave-one-adsorbate-out cross-validation.
\end{itemize}

The research agent is not given the final protocol as a predetermined workflow.
Instead, the protocol evolves during the research trajectory, subject to the
available skills, computational resources, and validation procedures.

\subsection{Execution budget}

The run is constrained by:

\begin{itemize}
    \item maximum wall-clock time: no hard cap; result production ran continuously for
    roughly five days within a multi-week trajectory;
    \item maximum model calls or token budget: a monthly token budget on the auditor LLM
    endpoint (a single-point-of-failure the trajectory had to route around);
    \item available CPUs/GPUs: a shared pool of H100 and V100 GPUs ($\sim$10 devices),
    with per-cell parallelism of two to four GPUs by system size;
    \item simulation concurrency: up to $\sim$16 parallel DFT relaxations across the V100
    pool (two per card), fewer for larger cells requiring density-grid splitting;
    \item filesystem/storage limits: a shared cluster filesystem plus node-local scratch
    for wavefunctions;
    \item explicit human-imposed constraints: DFT on H100/V100 only (never the slower GPU
    class), a no-kill policy on shared GPUs, and a no-delete-without-backup policy on
    result files.
\end{itemize}

\subsection{Trajectory statistics}

\begin{table}[ht]
    \centering
    \small
    \caption{Detailed trajectory statistics.}
    \label{tab:trajectory-detailed}
    \begin{tabular}{p{0.39\linewidth}p{0.54\linewidth}}
        \toprule
        \textbf{Statistic} & \textbf{Value} \\
        \midrule
        Total wall-clock time & $\sim$5 d DFT (multi-week trajectory) \\
        Research-agent invocations & $\sim$$10^4$ tool-calling turns (transcript) \\
        Auditor invocations & 6 material + standing recompute \\
        Plan revisions & 6 \\
        Explicit goal revisions & 2 \\
        Total tool calls & 13{,}679 (Bash 8.6k, Edit 2.3k, Read 1.1k, \ldots) \\
        Total skill calls & 8 skill types, invoked throughout \\
        Simulation runs & 50 production DFT + 8 clean + 171 MLIP \\
        Successful simulations & 21 converged + 29 best-effort (of 50) \\
        Failed simulations & several (resume loops, launch faults) \\
        Scientifically rejected runs & 8 (reconstruction blow-ups) \\
        Automatically recovered failures & $\ge$4 result-level + infra \\
        Human interventions & \rev{16} (Table~\ref{tab:human-ledger}) \\
        Generated figures & 8 manuscript figures + dashboards \\
        Final validated claims & 3 \\
        \bottomrule
    \end{tabular}
\end{table}

\newpage
\section{Research-Trajectory Analysis}
\label{app:trajectory-details}

\subsection{Trajectory overview}

\begin{figure}[ht]
    \centering
    \includegraphics[width=\linewidth]{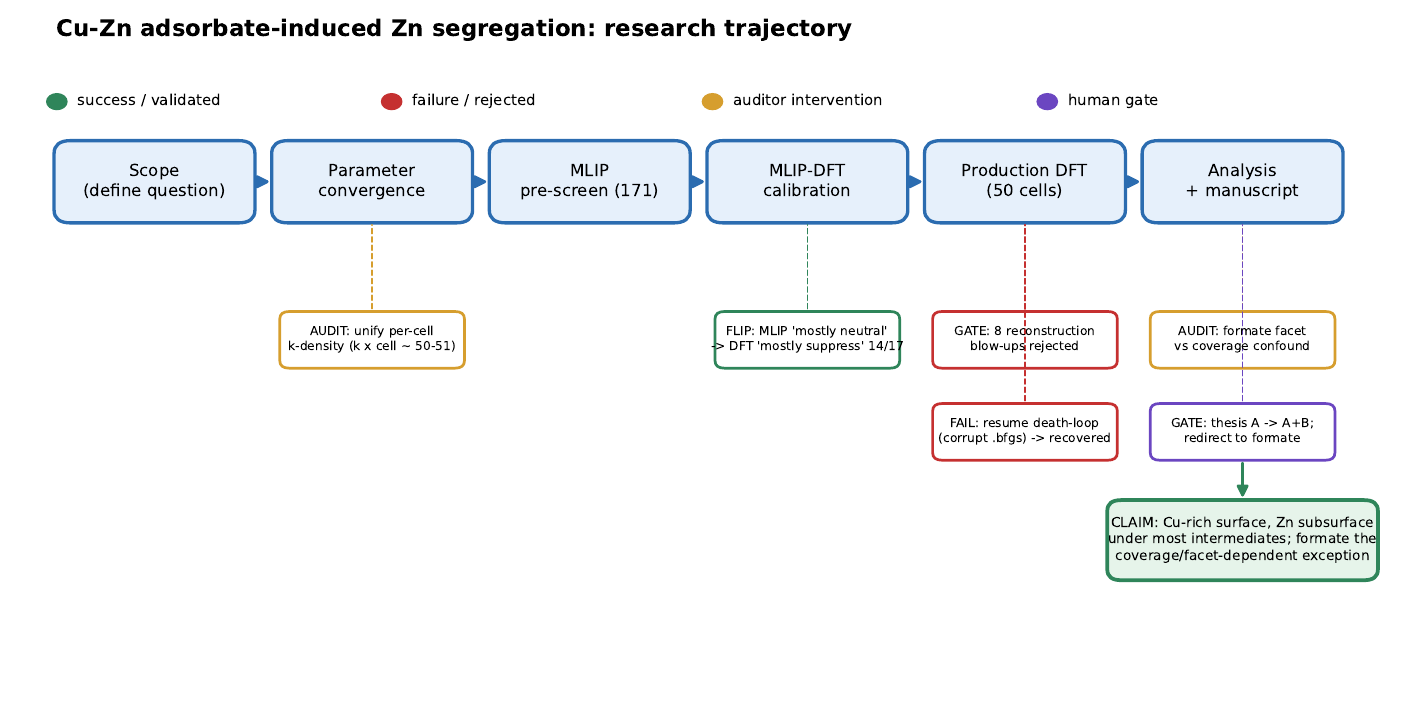}
    \caption{Complete research trajectory for the alloy-segregation case study.
    Nodes denote major scientific actions or state transitions rather
    than every low-level tool call.}
    \label{fig:trajectory}
\end{figure}

Figure~\ref{fig:trajectory} summarizes the complete research trajectory. Each major
experiment or decision is a node, with edges indicating the dependencies between
hypotheses, simulations, analyses, and plan revisions, and nodes are distinguished as
successful experiments, failed or rejected experiments, auditor-triggered corrections,
human gates, or the accepted scientific conclusion.

\subsection{Representative research transition I: hypothesis revision}

\paragraph{Initial state.}
From the machine-learning-potential pre-screen of 171 configurations, the agent's working
interpretation was that most CO$_2$-hydrogenation intermediates leave Zn segregation
roughly unchanged, with only scattered promotion---plausible because the surrogate
predicted small, mixed-sign shifts.

\paragraph{Experiment.}
The converged-$k$ DFT calculations that the surrogate had prioritised were executed as
paired segregated/substitutional cells, yielding $\Delta\Delta E_{\mathrm{seg}}$ for
each (facet, intermediate, coverage) point.

\paragraph{Observation.}
The DFT reversed the surrogate picture: on the 17-point science set the shift is
predominantly positive---14 suppress, 2 promote, 1 neutral---and the surrogate had
systematically mis-signed oxygen-bearing adsorbates and manufactured spurious high-coverage
blow-ups that DFT showed to be ordinary.

\paragraph{Mental-model update.}
The interpretation changed from ``intermediates are mostly neutral'' to ``most intermediates
\emph{suppress} segregation and drive Zn subsurface,'' and the surrogate was demoted from a
potential DFT stand-in to a queue-ordering tool only.

\paragraph{Plan update.}
This result became the paper's thesis; a facet-resolved analysis was added, which in turn
exposed that the lone strong promoter (formate) changes sign between facets and motivated a
dedicated coverage series (Transition~III).

This example \rev{demonstrates} that the research trajectory is evidence-dependent rather than a predefined workflow.

\subsection{Representative research transition II: failure recovery}

\paragraph{Failed action.}
A DFT relaxation restarted from a checkpoint entered a non-converging death-loop, repeatedly
resetting to a low ionic step and never reaching convergence while continuing to consume the
GPU---a failure invisible to a naive ``GPU busy'' health check.

\paragraph{Diagnosis.}
MatHarness traced the loop to a BFGS-history file truncated when a previous process was
killed mid-write: on restart the truncated history triggered a read-past-end error and an
immediate exit, which the scheduler then retried indefinitely.

\paragraph{State update.}
The unsuccessful attempt was retained in the persistent research state together
with the run logs, the underlying I/O error signature, and the diagnosis that the density
and wavefunction checkpoint were intact while only the optimizer history was corrupt.

\paragraph{Recovery.}
The corrupted BFGS history was discarded while the density/wavefunction checkpoint was kept,
and the run was restarted from atomic wavefunctions; the underlying scientific hypothesis did
not change. The fix was generalised into a reusable resume rule, converting a one-time human
diagnosis into standing autonomous-recovery behaviour.

This example is useful for distinguishing execution recovery from scientific
replanning.

\subsection{Representative research transition III: live-auditor intervention}

\paragraph{Pre-audit trajectory.}
The manuscript's centerpiece claim was that facet geometry inverts the sign of the formate
effect---promotion on the flat Cu(100) ($-0.43$~eV) but suppression on the stepped
Cu(211) ($+0.14$~eV).

\paragraph{Auditor finding.}
The two supporting points differ in \emph{both} facet and coverage ($\theta = 0.19$ vs
$0.08$), so the sign change is confounded and cannot be attributed to facet geometry alone---an
over-attributed conclusion resting on an uncontrolled comparison.

\paragraph{Corrective action.}
A targeted formate coverage series was planned (formate on Cu(100) and Cu(211) at matched
coverage, by removing molecules from the existing relaxed cells) to separate the facet effect
from the coverage effect; concurrently the quantitative phase-diagram claims, whose
lattice coupling was fixed by only two coverages per facet, were downgraded to qualitative.

\paragraph{Outcome.}
The supplementary coverage series \emph{refuted} the facet attribution, correcting an
over-claim before publication. Formate promotes on Cu(100) at every coverage tested
($\Delta\Delta E_{\mathrm{seg}} = -0.65, -0.11, -0.43$~eV at $\theta \approx 0.06, 0.13, 0.19$);
on Cu(211) it promotes when dilute ($-0.12$~eV at $\theta \approx 0.04$) and reverses to
suppression only at higher coverage ($+0.14$~eV at $\theta \approx 0.08$). At \emph{matched}
dilute coverage both facets therefore \emph{promote}, so the original ``facet flips the sign''
claim was a facet--coverage confound; facet geometry instead sets the coverage at which the sign
reverses. The auditor-triggered control thus overturned, rather than confirmed, the headline
attribution---the more valuable outcome, since it prevented a confounded claim from reaching print.

\subsection{Representative research transition IV: evolving research goal}

The initial objective \(G_0\) is to determine how CO$_2$-hydrogenation intermediates
reshape Zn segregation in Cu. After observing that both the \emph{strength} of suppression
and the \emph{sign} of the lone promoter track facet geometry (open and stepped faces
respond most strongly; formate flips sign between Cu(100) and Cu(211)), MatHarness modifies
the active research goal from

\begin{quote}
\texttt{which intermediates promote or suppress Zn segregation}
\end{quote}

to

\begin{quote}
\texttt{how intermediate chemistry and facet geometry jointly set the working-state Zn
location}.
\end{quote}

The transition is motivated by the facet-resolved DFT evidence and is subsequently retained
after auditor review, with the residual facet-versus-coverage confound flagged for the
supplementary confirmation described in Transition~III. This provides a concrete example of evidence-driven goal evolution
rather than uncontrolled research drift.

\newpage
\section{Scientific Claims and Evidence Traceability}
\label{app:claims}

To evaluate whether the final scientific conclusions remain grounded in the
research trajectory, we report the claim--evidence table below (Table~\ref{tab:claim-evidence}).

\begin{table}[ht]
    \centering
    \small
    \caption{Traceability of the final scientific claims to supporting evidence.}
    \label{tab:claim-evidence}
    \begin{tabular}{p{0.30\linewidth}p{0.31\linewidth}p{0.27\linewidth}}
        \toprule
        \textbf{Scientific claim} &
        \textbf{Supporting evidence} &
        \textbf{Validation} \\
        \midrule
Clean Zn segregates to all four facets, strongest on the open (211)/(110) faces &
        8 clean seg/sub DFT pairs; clean-$E_{\mathrm{seg}}$ values &
        Converged-$k$; independent MLIP cross-check \\

        Most CO$_2$-hydrogenation intermediates suppress segregation (14/17), strongest HCO/O on Cu(211) &
        25 seg/sub pairs $\to$ 17-point science set; Fig.~1--2 &
        Discernment gate; converged-$k$; MLIP MAE 0.21~eV, sign 13/15 \\

        Formate is the lone promoter; its sign reverses with coverage, and facet geometry sets the coverage at which it reverses (promotes on Cu(100) at all coverages tested; on Cu(211) promotes when dilute and suppresses at higher coverage); not descriptor-reducible &
        Matched-coverage formate series, $\Delta\Delta E_{\mathrm{seg}}$ per (facet, $\theta$); Appendix~\ref{app:trajectory-details}, Transition~III  &
        Auditor-triggered matched-coverage control; leave-one-adsorbate-out CV ($R^2<0$); coverage-series supplement \\
        \bottomrule
    \end{tabular}
\end{table}

This table provides a compact check that claims appearing in the generated
manuscript are not supported only by language-model reasoning but correspond to
identifiable computational evidence.

\newpage
\section{Provenance and Reproducibility}
\label{app:reproducibility}

\subsection{\rev{Data provenance}}
\label{app:data-provenance}

\begin{revblock}
All DFT calculations reported in this paper and in the generated manuscript (Appendix~\ref{app:generated-paper}), at both the Cu-rich and the Zn-rich composition limits, were planned, launched, monitored, analyzed, and audited by MatHarness (Table~\ref{tab:provenance}). No DFT data computed outside this trajectory, whether by the authors, by collaborators, or taken from the literature, enter any reported result.

The generated manuscript refers to the Cu-rich data set, and to the NequIP potential used for Cu-rich prescreening, as a ``companion study.'' This wording is an artifact of the second goal revision (the two-paper split; Table~\ref{tab:human-ledger}), after which the Writer agent drafted the Cu-rich and Zn-rich analyses as separate manuscripts before they were merged. It does not refer to an external study.

\begin{table}[ht]
    \centering
    \small
    \setlength{\tabcolsep}{4pt}
    \caption{Data provenance. Every entry was produced by MatHarness within the reported trajectory.}
    \label{tab:provenance}
    \begin{tabular}{p{0.2\linewidth}p{0.4\linewidth}p{0.32\linewidth}}
        \toprule
        \textbf{Data set} & \textbf{Calculations} & \textbf{Status and use} \\
        \midrule
        Cu-rich, clean & 8 cells (4 fcc facets $\times$ surface/bulk Zn) & Converged $k$; clean $E_{\mathrm{seg}}$ baseline \\
        Cu-rich, adsorbates & 25 surface/bulk pairs (50 relaxations); 8 removed by the reconstruction gate $\to$ 17-point science set & 21/50 force-converged, 29 energy-plateau best-effort; 14/17 suppression claim \\
        Formate coverage series & Matched-coverage cells on Cu(100) and Cu(211) & Auditor-triggered control; overturned the facet attribution \\
        Zn-rich, clean and CO & Surface/bulk Cu pairs on 4 hcp facets, clean and with one CO & Converged DFT; CO-induced reversal \\
        Electronic structure & PDOS, Bader charges, $\Delta\rho$ for CO on Cu-doped Zn(0001) & Converged DFT; Cu--CO mechanism \\
        MLIP screen & 171 NequIP evaluations; MACE/CHGNet cross-checks & Surrogate only; queue ordering and trend checks \\
        \bottomrule
    \end{tabular}
\end{table}
\end{revblock}

\subsection{Software and model versions}

\begin{table}[ht]
    \centering
    \small
    \caption{Software and model versions used in the case study.}
    \label{tab:software}
    \begin{tabular}{ll}
        \toprule
        \textbf{Component} & \textbf{Version / identifier} \\
        \midrule
        Roundtable Seats & claude-opus-4-8, gemini-3.1-pro-preview, gpt-5.5, deepseek-v4-pro \\
        Roundtable Chair & deepseek-v4-flash \\
        Idea Screening & gpt-4o-mini \\
        Research Agent & claude-opus-4-8 \\
        Auditor Agent & claude-haiku-4-5 (triage), claude-opus-4-8 (deep audit) \\
        Analyzer Agent & claude-opus-4-8 \\
        Writer Agent & claude-sonnet-4-5 \\
        MatHarness code & \url{https://github.com/PrentisAI/pippin} \\
        Python & 3.11 \\
        ASE / structure library & ASE 3.29.0 \\
        Simulation engine & Quantum ESPRESSO 7.3.1 (GPU) / 7.5 \\
        Interatomic potential & NequIP (pre-screen); MACE 0.3.16; CHGNet \\
        Analysis dependencies & NumPy, pandas, Matplotlib, scikit-learn \\
        \bottomrule
    \end{tabular}
\end{table}

\subsection{Third-party assets, attribution, and licenses}
\label{app:asset-licenses}

We use third-party scientific software, pseudopotentials, and hosted foundation
models under their upstream licenses or service terms. The principal directly
used assets are summarized in Table~\ref{tab:asset-licenses}. The corresponding
scientific software/model papers are cited in the manuscript where applicable.
We do not redistribute third-party source code, pretrained model weights, or
pseudopotential files with this submission; the table records the terms under
which the assets were used and the relevant upstream attribution.

\begin{table}[ht]
    \centering
    \small
    \caption{Principal third-party assets used in the reported study. License
    names refer to the upstream project distributions; hosted foundation models
    are used under the applicable provider/API terms rather than redistributed.}
    \label{tab:asset-licenses}
    \setlength{\tabcolsep}{3pt}
    \begin{tabular}{p{0.17\linewidth}p{0.21\linewidth}p{0.16\linewidth}p{0.30\linewidth}}
        \toprule
        \textbf{Asset} & \textbf{Role / identifier} & \textbf{License terms} & \textbf{Compliance} \\
        \midrule
        Quantum ESPRESSO & DFT; v7.3.1 (GPU) / 7.5 & GNU GPL & Used as installed software, not redistributed. \\
        ASE & structure construction; v3.29.0 & GNU LGPL & Used as an installed Python package. \\
        GBRV library & ultrasoft pseudopotentials & GNU GPL & Pseudopotential files are not redistributed. \\
        NequIP & MLIP pre-screen & MIT & Code/weights are not redistributed. \\
        MACE & independent MLIP cross-check; code v0.3.16 & MIT (code) & The exact checkpoint identifier is retained in run provenance and no weights are redistributed. \\
        CHGNet & independent cross-check & Modified BSD & Code/weights are not redistributed. \\
        NumPy / pandas & numerical analysis / tabulation & BSD 3-Clause & Used as installed Python packages; no source is redistributed. \\
        scikit-learn & descriptor regression / cross-validation & BSD 3-Clause & Used as an installed Python package; no source is redistributed. \\
        Matplotlib & scientific visualization & PSF-based, BSD-compatible license & Used as an installed Python package; no source is redistributed. \\
        Hosted foundation models & multi-agent roundtable, planning, Researcher, Auditor, Analyzer, Writer & provider/API terms & Accessed through authorized hosted endpoints; model weights are not redistributed. Exact deployed model identifiers are reported. \\
        \bottomrule
    \end{tabular}
\end{table}

The upstream project documentation identifies Quantum ESPRESSO as GNU-GPL
software, ASE as GNU-LGPL software, NequIP and the MACE code as MIT licensed,
CHGNet as Modified BSD, and the GBRV pseudopotential library as released under
the GNU public license. NumPy, pandas, and scikit-learn use BSD-style licenses,
while Matplotlib uses a PSF-based BSD-compatible license. License metadata was
checked against the corresponding upstream project documentation and LICENSE
files rather than inferred from package names. We use these assets
without modifying or repackaging their distributions. Where an upstream model
checkpoint can carry terms distinct from its software implementation (notably
MACE), the checkpoint is treated as a separate asset and its identifier/terms
are retained in provenance rather than inferred from the code license.

\subsection{Cost reporting}

To characterize the practical cost of long-horizon autoresearch, we report both
AI-inference and scientific-computation cost where instrumented:

\begin{table}[ht]
    \centering
    \small
    \caption{Compute and inference cost.}
    \label{tab:cost}
    \begin{tabular}{p{0.39\linewidth}p{0.54\linewidth}}
        \toprule
        \textbf{Cost measure} & \textbf{Value} \\
        \midrule
        Literature access & free arXiv retrieval in the idea screening stage \\
        CPU hours & negligible (compute was GPU-DFT) \\
        GPU hours & $\sim$970 (all 191 DFT cell-runs incl.\ convergence tests and restarts; 58 are the reported production + clean cells; wall-time $\times$ cards) \\
        Simulation wall-clock time & $\sim$5 d continuous DFT \\
        Total wall-clock duration & multi-week trajectory \\
        LLM stages & Costs span seven stages—ideation, idea screening, planning, execution, auditing, analysis, and writing. Centralized aggregation of total token usage is omitted due to a mixed payment model utilizing both fixed monthly subscriptions (Claude) and pay-as-you-go APIs (OpenRouter). \\
        Estimated LLM API cost & $\ll$ GPU cost \\
        \bottomrule
    \end{tabular}
\end{table}

The MatHarness pipeline incurs LLM-inference cost across \rev{its stages, including} study design, the planning
round-table, per-step live auditing (a cheap triage model on every step and an expensive
deep-audit model on flagged steps), novelty / prior-art assessment (the writing agent's
literature-gap check against arXiv), manuscript writing, and review---in addition to the
GPU-bound DFT compute. Because no central token ledger was kept, we \crfix{bound the LLM cost
using measured \emph{per-call unit costs} (live third-party API measurements)} rather than
report a single trajectory total. The important observation is scale: the
most expensive single call (a deep opus audit) is $\sim$\$$6\times10^{-3}$, so even
aggregated over the full trajectory---hundreds of triage/audit calls, several round-tables,
novelty checks, and writing passes---the LLM cost is at most single-digit to low-tens of
dollars, whereas the DFT campaign consumes order $10^2$--$10^3$ GPU-hours. For long-horizon
computational autoresearch of this kind, the binding cost is scientific compute rather than LLM tokens.

\clearpage
\section{Additional Evaluation}
\label{app:evaluation}
\subsection{Ablation Studies}

To evaluate the core mechanisms of \textsc{MatHarness}, we conducted controlled experiments using \texttt{claude-opus-4-8}. First, we compared a raw, toolless LLM (the ``no-harness'' baseline, lacking auditing, skills, and persistent state) against our full harness equipped with independent recomputation. Tested on an eight-material benchmark (four valid cases and four with seeded flaws), the raw LLM passed 0/4 valid cases and detected only 2/4 flawed ones. In contrast, the full harness correctly verified all 4/4 valid cases without false positives and caught all 4/4 flawed ones.

A second ablation evaluated our persistent mental model. By utilizing a persistent mental model carrying historical failure lessons, true-flaw recall on a lighter base model rose from 10/16 to 15/16, while a regression gate successfully prevented false kills. 

In addition to these quantitative benchmarks, our full-trajectory execution provides clear qualitative evidence for each underlying mechanism. Specifically, \textbf{persistent research state} preserved converged parameters and failure logs across context truncations and model restarts over a multi-week timeline; a \textbf{mental world model} enabled the agent to pivot when DFT contradicted initial MLIP surrogate predictions (Transition~I); \textbf{domain skills} supplied standardized convergence and validation gates without ad-hoc scripting; \textbf{goal evolution} allowed the research thesis to adapt from an intermediate-only scope to a facet-coverage analysis (Transition~IV); and \textbf{explicit planning} converted a 171-configuration space into a targeted boundary DFT queue rather than a naive grid search.

\subsection{Human Expert \rev{Reference} and Fixed Workflow\rev{s}}

We refer to conventional domain practice in our collaborating physical-chemistry laboratory as a human expert baseline. It is estimated that a project of this scope—encompassing structural validation, calculation monitoring, failure diagnosis, and iterative research goal adjustments—typically requires 1--2 months of dedicated human effort of a senior PhD while MatHarness accomplishes this in approximately 2 weeks. \rev{This is a collaborator-informed expert estimate, not a controlled human-subject timing study, and we do not claim a controlled speedup over human researchers.}

Static workflows effectively automate routine, pre-specified computational pipelines. However, they are fundamentally ill-suited for anticipating emergent scientific anomalies or navigating open-ended research trajectories. Throughout the case study, key decisions—such as overriding MLIP screening upon DFT convergence, correcting cross-facet $k$-point mismatches, and introducing matched-coverage controls—emerged dynamically from intermediate data analysis. Managing such non-linear, adaptive trajectories remains inherently beyond the reach of static workflows.

\newpage
\section{Prompts for Agents}
\label{app:prompts}
\subsection{Multi-Agent Ideation Roundtable Instruction}
\label{app:multi-agent}
\begin{quote}
\small
\textbf{Persona Initialization and Roundtable Discipline} \emph{Excerpt from the roundtable seat system prompt:} ``You are seat \texttt{\detokenize{seat.model_id}} in a multi-model human-agent round-table\ldots Evidence discipline: every claim must be grounded in one of---(1) experiment results in the briefing: cite the \texttt{\detokenize{idea_id}} and the specific finding; (2) mental-model beliefs: cite the belief statement; (3) published literature: provide the DOI or URL. Anything you assert beyond these three sources must be explicitly flagged \texttt{'(my claim, unverified)'}\ldots You're welcome to bring relevant knowledge, methods and ideas from adjacent fields---or even distant but structurally related ones---to bear on this question.''

\emph{In summary,} the initialization phase \rev{enforces} strict grounding: agents are prohibited from making unbacked assertions during brainstorming, while dynamic multi-agent panel composition introduces \rev{heterogeneous} agent personas.
\end{quote}

\begin{quote}
\small
\textbf{Multi-Turn Deliberative Debate and Convergence} \emph{Excerpt from the multi-round interaction instructions:} ``ROUND 1---your opening statement. Go through the other members ONE BY ONE, and within each member review EVERY numbered proposal in their report---do not skip any item. For each item give a reasoned verdict: agreement or critical suggestions, referencing items as e.g.\ \texttt{p02 \#3}\ldots ROUND $k$---address the points raised against positions you took, react to arguments that changed or sharpened your view (say explicitly if you updated, and why), and push unresolved disagreements forward with NEW reasoning or evidence---do not repeat yourself.''

\emph{In summary,} deliberative debate avoids superficial consensus by forcing exhaustive, item-by-item peer audits in initial rounds and requiring new evidence for updates in subsequent turns. An independent moderator monitors transcript stability across parallel rounds and terminates discussion once positions stabilize and argument density declines.
\end{quote}

\begin{quote}
\small
\textbf{Roundtable Chair Responsible for Dynamic Routing and Convergence Determination} \emph{Excerpt from the roundtable chair system prompt:} ``You are a terse meeting moderator judging convergence\ldots Judge whether it has CONVERGED: positions are stable, the latest round added no substantially new arguments or evidence, and further rounds would mostly repeat. Return ONLY JSON: \texttt{{"converged": true|false, "reason": "..."}}\ldots [For routing mode:] Decide which seats should speak NEXT. A seat should speak if it has a materially new point or a substantive (dis)agreement\ldots Reply with a comma-separated list of \texttt{seat\_ids} in speaking order, or \texttt{'none'}.''

\emph{In summary,} the roundtable chair optimizes discussion dynamics by assessing convergence for early termination and routing speaker turns based on argument novelty, eliminating conversational redundancy and minimizing extraneous LLM invocations.
\end{quote}

\subsection{Planning Stage Instruction}

\begin{quote}
\small
\emph{Excerpt from the planning stage system prompt:} ``You decompose an empirical research topic into primitives for experiment design. In particular, identify critical system parameters and domain-specific factors; if a property varies across configurations, environments, or conditions, it must be treated as an explicit variable rather than held constant. \ldots For each intended experiment, instantiate it into concrete, executable experimental parameters (specific variable names, operational ranges, target observables, and evaluation thresholds).''

\emph{In summary,} the planning stage produces a concrete and  executable experiment plan through adaptive experimental design.
\end{quote}

\subsection{Researcher Agent Instruction}

\begin{quote}
\small
\emph{Excerpt from the Researcher Agent system prompt:} ``You are an
autonomous research agent working in a sandboxed workspace. There is no human available to
answer questions mid-task; make your best judgment and keep going. Do not ask for files or
permission---investigate first. \ldots Any command that may run longer than $\sim$30\,s
\ldots MUST be launched in the BACKGROUND, then END YOUR TURN and wait for the completion
notification. \ldots A supervisor may occasionally inject a message prefixed with
[SUPERVISOR]. Treat it as authoritative senior guidance: read it, adjust course, and briefly
acknowledge what you changed.'' 

\emph{In summary,} the researcher agent is instructed to pursue an open-ended scientific objective by proposing and executing computational experiments rather than following a fixed workflow; to maintain a persistent research state and a separate mental world model; to validate every computational setup (convergence, consistent references, discernment gates) before drawing
conclusions; to keep hypotheses, observations, and accepted claims distinct; and to
synthesize the validated evidence into a manuscript, deferring to the auditor and to human gates where a conclusion cannot be justified autonomously.
\end{quote}

\subsection{Auditor Agent Instruction}

\begin{quote}
\small
\emph{Excerpt from the Auditor Agent system prompt:} ``You are an adversarial
research supervisor auditing an AI agent. Assume it may be cutting corners or
reporting broken results; prove whether it is. You judge on TWO axes: (A) PROCESS
INTEGRITY---is the agent doing the ACTUAL task, using required methods/tools,
running the REQUIRED ANALYSES (not a related-but-different one), not
faking/shortcutting? Inspect FILE CONTENT and the verbatim TRAJECTORY \ldots not
just narration. (B) RESULT VALIDITY---are the agent's ACTUAL results trustworthy
by the standards of this field, or self-evidently broken? \ldots FAIRNESS RULE
(critical): you do NOT have the target paper or its numbers \ldots NEVER flag a
result merely for differing from a specific target value \ldots Only flag results
that are self-evidently INVALID.''

\emph{In summary,} the auditor agent checks goal, plan, \crfix{execution,} evidence, and epistemic
consistency, emitting one of: continue, validate, replan, reframe, reject, recover, or
escalate. When necessary, it independently re-derives key conclusions with a different computational engine (e.g., a second interatomic potential, and where feasible a DFT
cross-check). Its independence is methodological rather than a second prompt over the same
model.
\end{quote}

\subsection{Analyzer Agent Instruction}

\begin{quote}
\small
\emph{Excerpt from the Analyzer Agent system prompt:} ``You are a rigorous analyzer. Your responsibility is to critically evaluate experimental results, detect performance anomalies, perform root-cause analysis, and extract scientific insights. You operate with absolute statistical rigor, strictly avoiding confirmation bias, hallucinated metrics, or unsupported claims. You need to read experiment Execution logs and output artifacts \ldots Always ground every assertion in concrete numerical evidence from the logs. Never assume an experiment succeeded just because it executed without runtime errors.''

\emph{In summary,} the analyzer serves as the rigorous data scientist, tasked with translating raw execution logs and experimental outputs into actionable scientific insights. It operates by auditing data integrity to filter out silent failures, data leakage, and random noise, quantitatively comparing current results against baseline metrics when applicable.
\end{quote}

\subsection{Writer Agent Instruction}

\begin{quote}
\small
\emph{Excerpt from the Writing Agent system prompt:} ``You are
the corresponding author, writing a Markdown manuscript draft. Use ONLY the
provided evidence numbers and plan parameters; do not fabricate; every result
sentence must trace back to evidence; write the methods section per the evidence's
ACTUAL method---MLIP values must NOT be written as DFT. \ldots State the mechanistic
story your current evidence can support first \ldots list limitations concisely in a
single closing section rather than repeating disclaimers throughout; a genuine gap
is noted in one line and deferred to the follow-up plan (\texttt{paper\_plan}).''
The paper-planning step is prompted separately: ``\ldots judge what specific
additional calculations are still needed to make the paper solid (missing
controls / robustness / ablations / reviewer-mandatory data points); list only
what genuinely needs new computation, not pure writing edits.''

\emph{In summary,} the writer agent evaluates whether the current evidence is strong and complete enough for a meaningful scientific contribution. If gaps are found, it feeds them directly back to the next research cycle, initiating a new iteration to form hypotheses or plan missing experiments. The actual writing and internal review only begin when the writer agent confirms that both novelty and completeness requirements are fully satisfied.
\end{quote}

\newpage
\section{Example Research State Snapshot}
\label{app:state-snapshot}

The following schematic illustrates the type of information retained at an
intermediate point in the alloy-segregation research trajectory (an actual
snapshot from the run, lightly anonymized).

\begin{verbatim}
Initial objective:
  How do CO2-hydrogenation intermediates reshape Zn segregation in Cu?

Active goal:
  Characterize the sign and magnitude of ddE_seg across four facets.

Current plan:
  1. Complete converged-k DFT on the 25 seg/sub pairs.
  2. Apply the |ddE_seg| > 0.5 eV discernment gate.
  3. Fit a cluster expansion and run SGCMC for environmental coverage.

Working hypotheses:
  H1: Most intermediates suppress segregation (from DFT so far).
  H2: Open/stepped faces respond more strongly than close-packed terraces.

Validated evidence:
  E1: Clean Zn segregates on all four facets (-0.24 to -0.30 eV).
  E2: HCO/Cu(211) is the strongest suppressor (+0.40 eV).

Open uncertainties:
  U1: Is formate's promotion driven by facet or by coverage?
  U2: Are the eight high-coverage blow-ups real reconstructions?

Failed / rejected directions:
  F1: Eight high-coverage points rejected by the 0.5 eV gate.

Current scientific claims:
  C1 [tentative]: Facet geometry sets the sign of the effect.
  C2 [validated]: Most intermediates suppress segregation.

Next proposed action:
  Fit the cluster expansion and run SGCMC for the T-p phase diagram.

Rationale:
  Convert dilute-limit energetics into working-condition predictions.
\end{verbatim}

\section{Example Just-in-time Auditing}
\label{app:auditor-example}

The following is an actual auditor intervention from the case study.

\begin{verbatim}
Audit trigger:
  Promotion of a cross-facet segregation-energy comparison to a claim.

Active research goal:
  Compare ddE_seg across the four facets.

Agent's proposed conclusion/action:
  Report a facet ordering of segregation energies.

Relevant evidence:
  Per-facet DFT computed with per-facet k-point meshes.

Auditor finding:
  The k-point meshes differed across facets, so the energies were not
  directly comparable and the ordering could be an artifact of sampling.

Detected issue:
  missing control / inconsistent parameters

Required corrective action:
  Re-run with a unified per-cell k-point density (k x cell ~ 50-51).

Outcome:
  Energies made directly comparable; the facet ordering held under the
  consistent setting, and per-cell k-density became a standing gate.
\end{verbatim}

\section{Example Research Artifacts}
\label{app:artifacts}

Other researcher artifacts take the following form:

\begin{table}[ht]
    \centering
    \small
    \caption{Example research artifacts}
    \label{tab:artifact-manifest}
    \begin{tabular}{lll}
        \toprule
        \textbf{Artifact} & \textbf{Research step} & \textbf{Role} \\
        \midrule
        \texttt{experiment\_plan.json} & Planning & Active-learning experiment plan \\
        \texttt{novelty\_prescreen.json} & Novelty check & Literature novelty screen \\
        \texttt{plan\_audit.json} & Auditing & Auditor review of the plan \\
        \texttt{<cell>/result.json} & Production DFT & Per-cell converged-$k$ energetics \\
        \texttt{analysis\_master.json} & Aggregation & Paired $\Delta\Delta E_{\mathrm{seg}}$ + gate vetting \\
        \texttt{F1\_seg\_clean\_vs\_ads.png} & Final evidence & Adsorbate-induced shift \\
        \bottomrule
    \end{tabular}
\end{table}

\newpage
\section{Generated Alloy-Segregation Manuscript}
\label{app:generated-paper}

\begin{quote}
\small
\textbf{Disclaimer:} The manuscript presented in this section, titled \emph{``Adsorbate-Induced Surface Segregation in Copper--Zinc Model Alloys for Methanol Synthesis''}, was autonomously synthesized by the MatHarness framework. While its underlying physical mechanism, computational energetics, and core scientific claims have been independently audited and validated by human domain experts, the text, formatting, and mathematical exposition are reproduced verbatim to demonstrate the agent's raw output. Consequently, stylistic anomalies, minor oversights, or formatting artifacts may remain. \rev{\textbf{Provenance note:} where the manuscript attributes the Cu-rich DFT data or the Cu-rich NequIP potential to a ``companion study,'' this refers to the Cu-rich draft that MatHarness itself produced after the two-paper split; all DFT in the manuscript was computed by MatHarness within the reported trajectory (Appendix~\ref{app:data-provenance}).}
\end{quote}

\clearpage 

\includepdf[
  pages=-, 
  pagecommand={\thispagestyle{plain}}, 
  scale=0.88
]{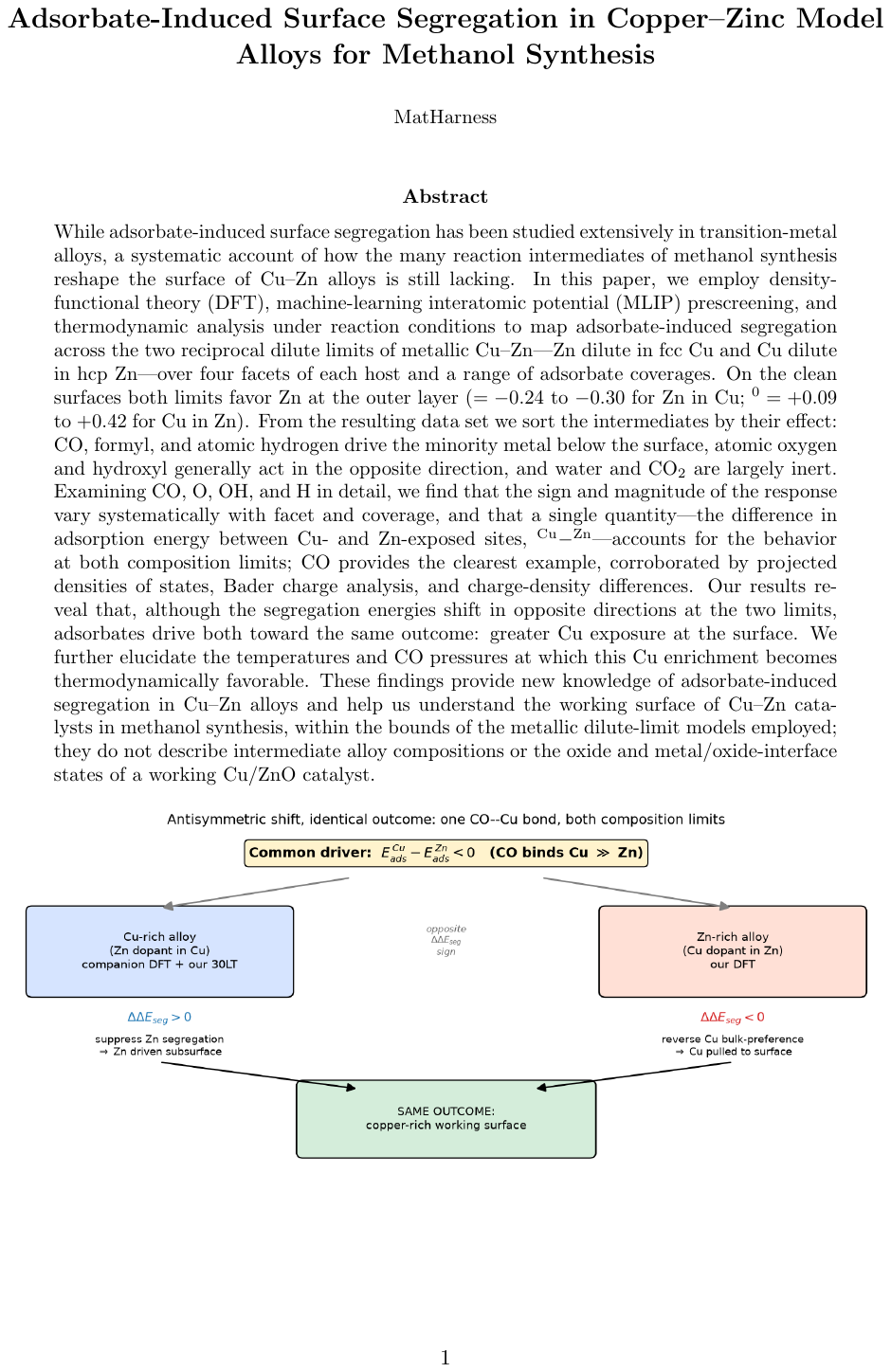}

\end{document}